\documentclass[journal]{IEEEtran}
\usepackage{graphicx}
\usepackage{cite}
\usepackage{picinpar}
\usepackage{amsmath}
\usepackage{url}
\usepackage{flushend}
\usepackage[latin1]{inputenc}
\usepackage{colortbl}
\usepackage{soul}
\usepackage{multirow}
\usepackage{pifont}
\usepackage{color}
\usepackage{alltt}
\usepackage[hidelinks]{hyperref}
\usepackage{enumerate}
\usepackage{siunitx}
\usepackage{breakurl}
\usepackage{epstopdf}
\usepackage{pbox}
\usepackage{pdfpages}
\usepackage{bm}

\usepackage{makecell}
\usepackage{stfloats}
\usepackage{textcomp}

\usepackage{float}
\usepackage{booktabs}
\usepackage{threeparttable}
\usepackage{footnote}

\begin{document}

\title{FlexDelta: A flexure-based fully decoupled parallel $xyz$ positioning stage with long stroke}

\author{
  Qianjun Zhang, 
  Wei Dong, \emph{Senior Member, IEEE},
  Qingsong Xu, \emph{Senior Member, IEEE},\\
  Bimal J. Goteea,
  and Yongzhuo Gao, \emph{Member, IEEE}

  \thanks{
    Research supported by National Natural Science Foundation of China under Grant No. 51475113 and 51521003, the State Key Lab of Self-planned Project under Grant No. SKLRS202004B, and the Programme of Introducing Talents of Discipline to Universities under Grant No. B07018. (\emph{Corresponding Author: Yongzhuo Gao.})

    Qianjun Zhang, Wei Dong, Bimal J. Goteea, and Yongzhuo Gao are with the State Key Laboratory of Robotics and System, Harbin Institute of Technology, Harbin 150000, China (email: zhang\_qj@stu.hit.edu.cn; dongwei@hit.edu.cn; preetam.goteea@hit.edu.cn; gaoyongzhuo@hit.edu.cn).

    Qingsong Xu is with the Department of Electromechanical Engineering, Faculty of Science and Technology, University of Macau, Macau 999078, China (e-mail: qsxu@umac.mo).
  }
}



\maketitle

\begin{abstract}
  Decoupled parallel $xyz$ positioning stages with large stroke have been desired in high-speed and precise positioning fields. However, currently such stages are either short in stroke or unqualified in parasitic motion and coupling rate. This paper proposes a novel flexure-based decoupled parallel $xyz$ positioning stage (FlexDelta) and conducts its conceptual design, modeling, and experimental study. Firstly, the working principle of FlexDelta is introduced, followed by its mechanism design with flexure.
  Secondly, the stiffness model of flexure is established via matrix-based Castigliano's second theorem, and the influence of its lateral stiffness on the stiffness model of FlexDelta is comprehensively investigated and then optimally designed. Finally, experimental study was carried out based on the prototype fabricated. The results reveal that the positioning stage features centimeter-stroke in three axes, with coupling rate less than 0.53\%, parasitic motion less than 1.72 mrad over full range. And its natural frequencies are 20.8 Hz, 20.8 Hz, and 22.4 Hz for $x$, $y$, and $z$ axis respectively. Multi-axis path tracking tests were also carried out, which validates its dynamic performance with micrometer error.
\end{abstract}


\section{Introduction}
\IEEEPARstart
{T}{ranslational} $xyz$ positioning stages with large stroke and high speed have been desired in various precise positioning fields such as 3D laser lithography \cite{lafratta2020making,serge2020motion,zhang2017large}, cell manipulation \cite{zhang2019robotic} and microscope scanning \cite{werner2010design,weckenmann2007long}. 

Currently, there are mainly two kinds of translational $xyz$ positioning stages available, i.e., serially stacked stages based on roller bearings and flexure guides-based positioning system. The former kind is a widely used scheme for its inexpensive cost and maturity. However, it suffers from friction, creep, backlash, and abrasion, thus its difficulty to gain high precision even under periodic maintenance \cite{bettahar20206, zhan2018error}. And its bulky structure makes it hard to obtain quick dynamic response. Otherwise, the latter kind leverage the elastic deformation of the flexure hinges which eliminate the nondeterministic effects mentioned above \cite{yang2022along,du2013piezo, chen2019pzt}. Besides, its compact size and light weight allows for designing high-dynamic $xyz$ stage freely. 

Further, flexure guides-based $xyz$ positioning stage can be mainly classified as coupled parallel $xyz$ positioning stage \cite{hesselbach2004performance,xie2021design,yun2011optimal} and decoupled one \cite{bacher2001delta3, tang2006large, li2010totally, hao2015design, hao20093, awtar2021experimental,zhang2018design}. The coupled one utilizes complicated kinematics to transfer the motions from each branch to output stage, whose precision relies a lot on the manufacturing and assembling accuracy that causes modeling uncertainty. And its dynamic coupling is usually intractable to consider for control when high speed and precision are needed. Differently, for decoupled one, the motions of branches directly transfer to the output stage motion, sparing the complicated kinematic modelling and dynamic coupling. And it's believed decoupled positioning stage has less, even no motion loss compared with coupled one. Therefore, it's more preferred for its straightforward and simple control algorithm, and easiness to realize high precision with less kinematic model uncertainty caused by geometry and assembly errors.

To this end, researchers have paid great attention to the development of decoupled parallel $xyz$ positioning stage based on flexure guides over the past two decades. Early, Bacher \cite{bacher2001delta3} conducted the design and experimental test of a flexure-based decoupled $xyz$ stage, which possesses workspace of 2$\times$2$\times$2 mm$^3$ and natural frequency of 8.5 Hz under dimension of 1 dm$^{3}$. Likewise, Tang \textit{et al.} \cite{tang2006large}, Xu \cite{li2010totally,zhang2018design}, and Hao \cite{hao2015design} also proposed their designs of decoupled $xyz$ positioning stage based on various kinds of flexure hinges. These designs all have working range about or less than 2$\times$2$\times$2 mm$^3$, and their parasitic motion and coupling rate are not comprehensively analysed. To further increase the stroke of flexure-based decoupled $xyz$ stage, Hao \cite{hao20093} implemented the design of a monolithic decoupled $xyz$ compliant parallel mechanism. Despite its relatively low natural frequency (9.0 Hz), the simulation results reveal a working space of 9.5$\times$9.5$\times$9.5 mm$^3$ with an envelope size of 300$\times$300$\times$300 mm$^3$. More recently, Awtar \textit{et al.} \cite{awtar2021experimental} presented the experimental work of a large-range parallel kinematic $xyz$ flexure mechanism and it has centimeter stroke over all three DOFs with dimension of 150$\times$150$\times$150 mm$^3$, which is the design so far owning the largest working space-to-envelope size ratio, i.e. 0.0296 $\%$, among the surveyed designs. However, this is at the cost of realtively large parasitic motion (9.5 mrad) and coupling rate (11.6 $\%$) over the full range, which hinders its further application.
There also exist some flexure-based decoupled parallel $xyz$ stages with submillimeter stroke specilized for ultra ultra-high precision scenarios \cite{chen2022design,chen2021design,lin2019decoupling,ling2018design,liu2018review,zhang20223}. However, these are beyond the scope of this work since we aim at long-stroke ones, typically above millimeter.

Apparently, it's challenging to feature all characteristics, i.e., large stroke, high dynamics, low coupling rate and parasitic motion, simultaneous for previous works. Large working space of current designs is at the cost of bulky dimension and therefore unsuitable for desktop application. And compromise must be made to obtain either larger stroke or higher natural frequency under limited dimension, which puts it into a dilemma because of the contradictory relationships between them when subjected to limited drive force and compact size. Most importantly, ongoing efforts are needed to eliminate the coupling rate and parasitic motion of parallel $xyz$ stage which otherwise decreases its precision.

To solve the problem above, this paper proposed a novel flexure-based fully decoupled parallel $xyz$ positioning stage, named FlexDelta hereinafter, with centimeter stroke, high dynamic performance, and the highest compactness ratio which is defined as ratio between working space to dimension size (0.0465\%), known to our best knowledge. And its parasitic motion and coupling rate are both significantly mitigated as well. 
The main contributions of this paper are twofold. First, a novel mechanical configuration of flexure-based decoupled parallel $xyz$  stage is proposed and experimentally verified, which features preferable performance compared with previous design. Secondly, the influence of lateral stiffness on the stiffness model of mechanism, ignored in former designs, is comprehensively investigated in this work. And it proves critical to mitigate coupling rate among each axis.

\section{Mechanism Design of the FlexDelta}
\subsection{Conceptual design and working principle}
\begin{figure}[htpb]
  \centering
  \includegraphics[width=8cm]{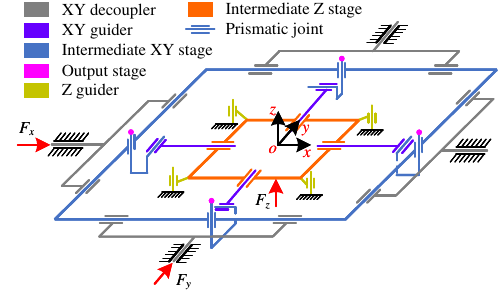}
  \caption{Schematic diagram and decoupled working principle of the proposed FlexDelta.}
  \label{Schematicdiagramanddecoupledworkingprinciple}
\end{figure}

The schematic diagram and working principle of the proposed FlexDelta is illustrated as Fig. \ref{Schematicdiagramanddecoupledworkingprinciple}. It mainly consists of XY decouplers, XY guider, Z guider, intermediate XY stage (with Z decouplers on it), intermediate Z stage, and output stage. All the kinematic pairs in this mechanism are prismatic joints.

The function of XY decouplers is to combinate with intermediate XY stage, to generate \textit{x} and \textit{y} decoupled in-plane motion under driving force $F_x$ and $F_y$. The driving force in \textit{z} direction is applied on the intermediate Z stage, which can only move along \textit{z} axis under the guiding of Z guiders.
To decouple the motion in \textit{z} and \textit{xy}, we propose a decoupling scheme as demonstrated in Fig. \ref{Schematicdiagramanddecoupledworkingprinciple}. The prismatic joints in the intermediate XY stage are arranged vertically and symmetrically. Meanwhile, the one end of XY guider is connected with the vertical prismatic joints (Z decouplers) in intermediate XY stage while the other end is fixed in intermediate Z stage. With this, the parts responsible for \textit{x} and \textit{y} motion will be able to move within \textit{xy} plane while those for \textit{z} motion can move along \textit{z} axis independently.

\subsection{Mechanism design with flexure}
\begin{figure}[htpb]
  \centering
  \includegraphics[width=8cm]{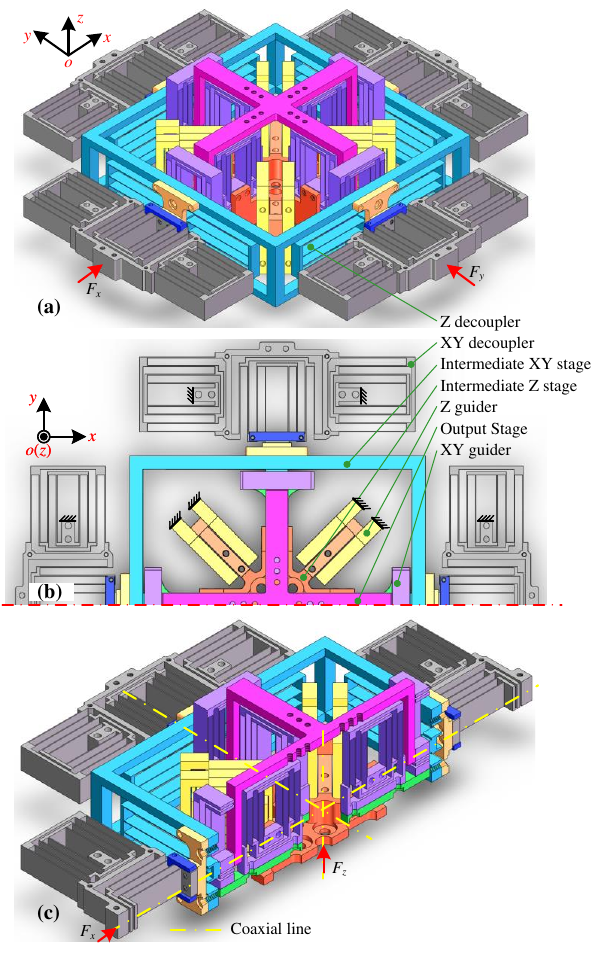}
  \caption{Design overview of the proposed FlexDelta. (a) Isometric view. (b) Half top view. (c) Section view. }
  \label{DesignOverviewOfTheProposedFlexDelta}
\end{figure}

\begin{figure}[htpb]
  \centering
  \includegraphics[width=8cm]{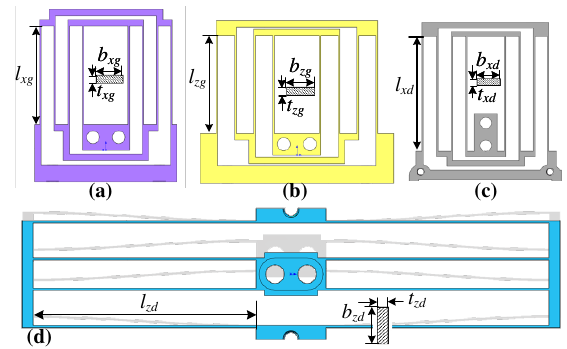}
  \caption{Four MCPFs functioning as prismatic joints in (a) XY guider, (b) Z guider, (c) XY decoupler, and (d) Z decoupler.}
  \label{FourMCPFsFunctioningAsPrismaticJoints}
\end{figure}

The conceptual design in Fig. \ref{Schematicdiagramanddecoupledworkingprinciple} is implemented with flexure as in Fig. \ref{DesignOverviewOfTheProposedFlexDelta}. All prismatic joints are realized with different MCPF units exhibited in Fig. \ref{FourMCPFsFunctioningAsPrismaticJoints}. Because MCPF is especially appropriate for designing translational freedom \cite{hao2015design,xu2011new}, whose stroke, motional stiffness (stiffness in motion direction), and lateral stiffness (stiffness in out-of-plane direction of MCPF) are changeable by simply adjusting its beam layers, width, thickness and length.

\begin{figure}[htpb]
  \centering
  \includegraphics[width=8cm]{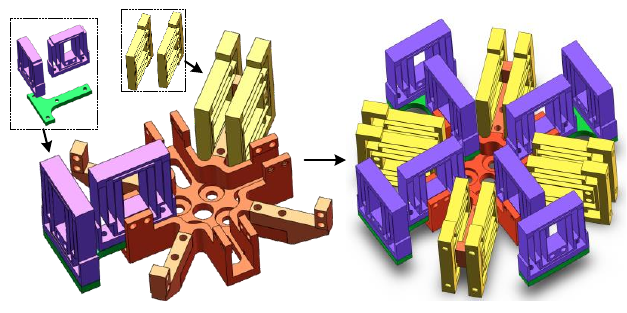}
  \caption{Decoupling mechanism of $z$ axis}
  \label{DecouplingMechanism}
\end{figure}

The XY decoupler is a sophisticated solution in our previous work \cite{xu2011new}, and we slice it into four blocks (XY decoupler) and connect them to intermediate XY stage.
The key point of FlexDelta is the decoupling mechnism for $z$-axis motion, which consists of Z guider, XY guider and intermediate Z stage as demonstrated in Fig.\ref{DecouplingMechanism}. Here, one pair of MCPFs are linked together perpendicularly then arranged in fours to build up XY guider, while eight MCPFs make up Z guider. The Z guider and XY guider are nested vertically in the intermediate XY stage, and the intermediate XY stage has Z decouplers direct on it for decoupling the $z$ motion. When $z$ axis moves, Z guider and Z decoupler deforms to provide motion, while all other flexures remains still due to large lateral or axial stiffness.
Likewise, when $x/y$ axis moves, the XY decoupler and XY guider deform, while Z guider and Z decoupler stay stationary. We refer readers to our supplementary video where visual demonstration is available. 
The assumption supporting this is the lateral and axial stiffness of an MCPF are significantly larger than its motional stiffness. Therefore, stiffness modeling and optimization design are essential to satisfy this assumption.

In early design period of FlexDelta, some primary methods are practical to reduce the parasitic motion and coupling rate:

1) Using MCPF as prismatic joint. As analyzed above, MCPF has smaller spanning because of its folded design. Simply increasing its layers will enlarge its total deformation, which diminishes the parasitic motion according to parallelogram principle. In addition, applying the force near the stiffness center, which is quite the occasion of an MCPF, helps to minimize the parasitic motion \cite{hao2015design}.

2) Employing symmetrical configuration. Asymmetric design inevitably induces parasitic motion, which increase with stroke, while for symmetry design such problem is avoidable. Hence, the proposed FlexDelta is configured symmetrically, even bisymmetrically for most modules.

3) Arranging parts of each axis coaxially. Eccentric force will necessarily induce tipping moment in the mechanism, which in turn causes undesired parasitic rotational motion. To avoid this condition, parts of each axis are arranged coaxially with driving force, displayed as Fig. \ref{DesignOverviewOfTheProposedFlexDelta}(c).

In conclusion, the main merit of this design lie in its fully decoupled parallel design, which guarantees homogeneous performance of each axis. And its nested, symmetric configuration and coaxial arrangement helps to mitigate parasitic motion and coupling rate even with long stroke.

\section{Stiffness Modelling}
To directly establish stiffness model of such complicated mechanism is nontrivial. However, the modelling can be vastly simplified if we model the four MCPFs first and then build up the stiffness of the whole mechanism. 

\subsection{Stiffness modelling of MCPF}
\begin{figure}[htpb]
  \centering
  \includegraphics[width=8cm]{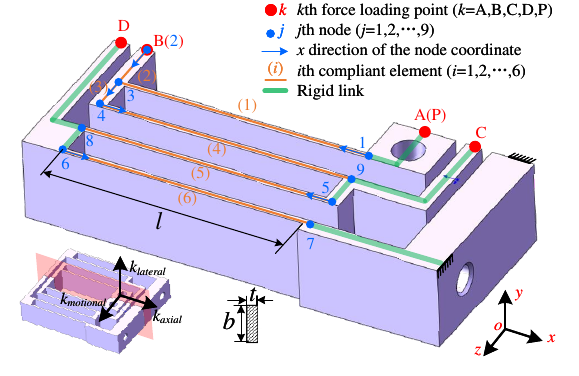}
  \caption{Skeleton representation of half an MCPF.}
  \label{SkeletonRepresentationOfHalfMCPF}
\end{figure}

For an MCPF functioning in FlexDelta, there are three types of stiffness, i.e. motional stiffness, lateral stiffness and axial stiffness. Their definition is illustrated as Fig. \ref{SkeletonRepresentationOfHalfMCPF}. Here, both its motional stiffness and lateral stiffness are considered. The reason why we ignore the axial stiffness will be discussed in Section V.
Basically, the motional stiffness determines the working range and natural frequency of the MCPF, hence the entire FlexDelta, while the lateral stiffness influences the stiffness model of each branch chain which will be clarified later in III.C. In previous work, the former is regularly done via simplified model \cite{xu2013design}, and the latter is generally ignored, which causes the model discrepancy. Here, we established the unified analytical model of both stiffness for a general MCPF via matrix-based Castigliano's second theorem. 

Half an MCPF is considered due to its symmetry, and its skeleton representation is illustrated as Fig. \ref{SkeletonRepresentationOfHalfMCPF}. The point P represents for the position where the driving force is loaded, and point A, B, C, D are where reaction force occurs. 

\begin{figure}[htpb]
  \centering
  \includegraphics[width=8cm]{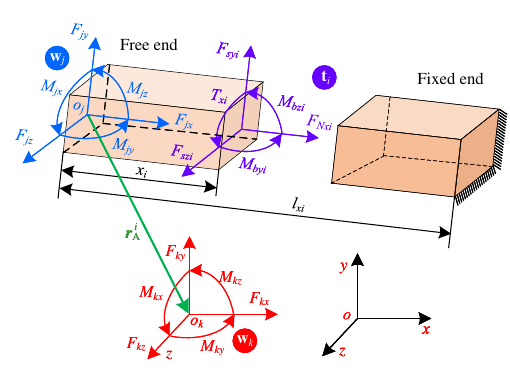}
  \caption{ Transformation from loading force $\boldsymbol w_k$ to internal force $\boldsymbol t_i$ of \textit{i}th compliant element.}
  \label{Transformation}
\end{figure}

To figure out the internal force $\mathbf {t_i} (i=1,2,...,6)$ of each compliant element, the loading force $\mathbf w_k (k=\rm{A, B, C, D, P})$ is first transferred to its free end, symbolized as $\mathbf w_j (j=i)$, then the internal forces $\mathbf t_i$ can be expressed in terms of $\mathbf w_j$. This transformation process and its coordinate definition is demonstrated as Fig. \ref{Transformation}.

The force transformation matrix from $o_k$ to $o_j$ is 
\begin{equation}
  \label{deqn_ex1}
  \mathbf{J}_{k}^{j}=\left[ \begin{matrix}
    \mathbf{R}_{k}^{j} & \mathbf{0}  \\
    \mathbf{s}\left( \mathbf{r}_{k}^{j} \right) & \mathbf{R}_{k}^{j}  \\
  \end{matrix} \right],
\end{equation}
where $\mathbf r_k^j$ is the location vector of $o_k$ with respect to $o_j$, $\mathbf R_k^j$ is the rotation matrix from $o_k$ to $o_j$, and $\mathbf s(\cdot)$ is the skew-symmetric operator.
Then, the internal force $\mathbf {t_i} (i=1,2,...,6)$ is calculated as
\begin{equation}
  \label{deqn_ex1}
  {{\mathbf{t}}_{i}}=\mathbf{R}_{wt}^{i}{{\mathbf{w}}_{j}},i=j=1,2,\ldots ,6,
\end{equation}
where $\mathbf R_{wt}^i$ is the matrix transferring $\mathbf w_j$ to $\mathbf t_i$.

Then the strain energy of the MCPF can be accumulated as 
\begin{equation}
  \label{deqn_ex1}
  U=\sum\limits_{i=1}^{7}{\int_{0}^{{{l}_{xi}}}\frac{1}{2}{\mathbf{C}{{\mathbf{o}}_{i}}\left( {{\mathbf{t}}_{i}}\odot {{\mathbf{t}}_{i}} \right)}}d{{x}_{i}},
\end{equation}
with $ \odot $ the Hadamard product, $\mathbf Co_i$ the coefficient vector expressed as
\begin{equation}
  \label{deqn_ex1}
  \mathbf{C}{{\mathbf{o}}_{i}}=[\frac{1}{E{{A}_{i}}} \frac{\alpha }{G{{A}_{i}}}  \frac{\alpha }{G{{A}_{i}}} \frac{1}{G{{I}_{pi}}}  \frac{1}{E{{I}_{yi}}}  \frac{1}{E{{I}_{zi}}}], i=1\sim 6.
\end{equation}
Here, $E$ is the Young's modulus, $G$ is the shear modulus, $\alpha$ is the shear factor (1.2 for rectangular cross-section), $A$ is the area of cross-section, $I_y$ and $I_z$ is are the inertial moment with respect to $y$ and $z$ axis respectively, and $I_{p}$ is the torsional moment of inertia which is interpolated as ${{I}_{p}}={{t}^{3}}b ({1}/{3}\;-0.21{t}/{b}\;+0.0175{{{t}^{5}}}/{{{b}^{5}}})$ for rectangular cross-section.

For the motional stiffness, the driving force ${{\mathbf{w}}_{P}}$ and the reaction force ${{\mathbf{w}}_{k}}$ induced are expressed as 
\begin{equation}
  \label{deqn_ex1}
  \left\{ \begin{aligned}
    & {{\mathbf{w}}_{P}}={{\left[ \begin{matrix}
     0 & 0 & {{F}_{Pz}} & 0 & 0 & 0  \\
  \end{matrix} \right]}^{\top }} \\ 
   & {{\mathbf{w}}_{k}}={{\left[ \begin{matrix}
     {{F}_{kx}} & {{F}_{ky}} & 0 & 0 & 0 & {{M}_{kz}}  \\
  \end{matrix} \right]}^{\top }},k=\rm{A,B,C,D} \\ 
  \end{aligned} \right.  
\end{equation}

For the lateral stiffness, the driving force ${{\mathbf{w}}_{P}}$  and the reaction force ${{\mathbf{w}}_{k}}$ induced are expressed as
\begin{equation}
  \label{deqn_ex3}
  \left\{ \begin{aligned}
    & {{\mathbf{w}}_{P}}={{\left[ \begin{matrix}
    0 & {{F}_{Py}} & 0 & 0 & 0 & 0  \\
  \end{matrix} \right]}^{\top }} \\ 
  & {{\mathbf{w}}_{k}}={{\left[ \begin{matrix}
    0 & 0 & {{F}_{kz}} & {{M}_{kx}} & {{M}_{ky}} & 0  \\
  \end{matrix} \right]}^{\top }}, k=\rm{A,B,C,D} \\ 
  \end{aligned} \right.
\end{equation}

With all the unknown reaction force solved, both the motional stiffness and lateral stiffness of MCPF can be worked out as
\begin{equation}
  \label{deqn_ex5}
  {{k}_{motional}}=\frac{1}{{\partial U}/{\partial {{F}_{Pz}}}\;}, {{k}_{lateral}}=\frac{1}{{\partial U}/{\partial {{F}_{Py}}}\;}
\end{equation}

We define the stiffness ratio $\eta = k_{lateral} / k_{motional}$ , and a larger $\eta$ is critical for the stiffness model of each branch chain, which will be analyzed later.  

\subsection{Model verification and parameter influence}
\begin{figure}[htpb]
  \centering
  \includegraphics[width=8cm]{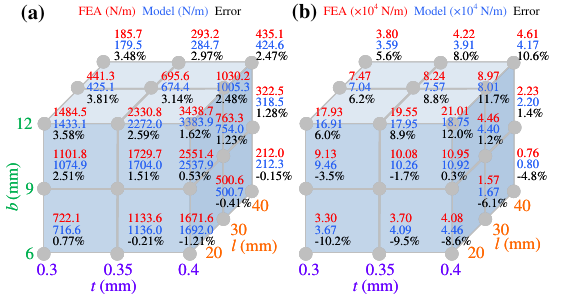}
  \caption{Model verification with FEA. (a) Motional stiffness. (b) Lateral stiffness.}
  \label{ModelVerificationWithFEA}
\end{figure}

To verify the accuracy of the established analytical model, FEA via ANSYS is utilized for comparison. The thickness $t$, length $l$, and width $b$ of the compliant element are selected as design parameters. We choose 3 points, i.e. minimum, median and maximum for each parameter. And to be intuitive, we verify the 19 points arranging in the surface of cabinet projection formed by $t, l, b$, as Fig. \ref{ModelVerificationWithFEA} displayed. Here, aluminum alloy is adopted as material, whose Young's modulus is $E$=71 GPa, shear modulus is $G$=26.7 GPa.

\begin{figure*}[htpb]
  \centering
  \includegraphics[width=16cm]{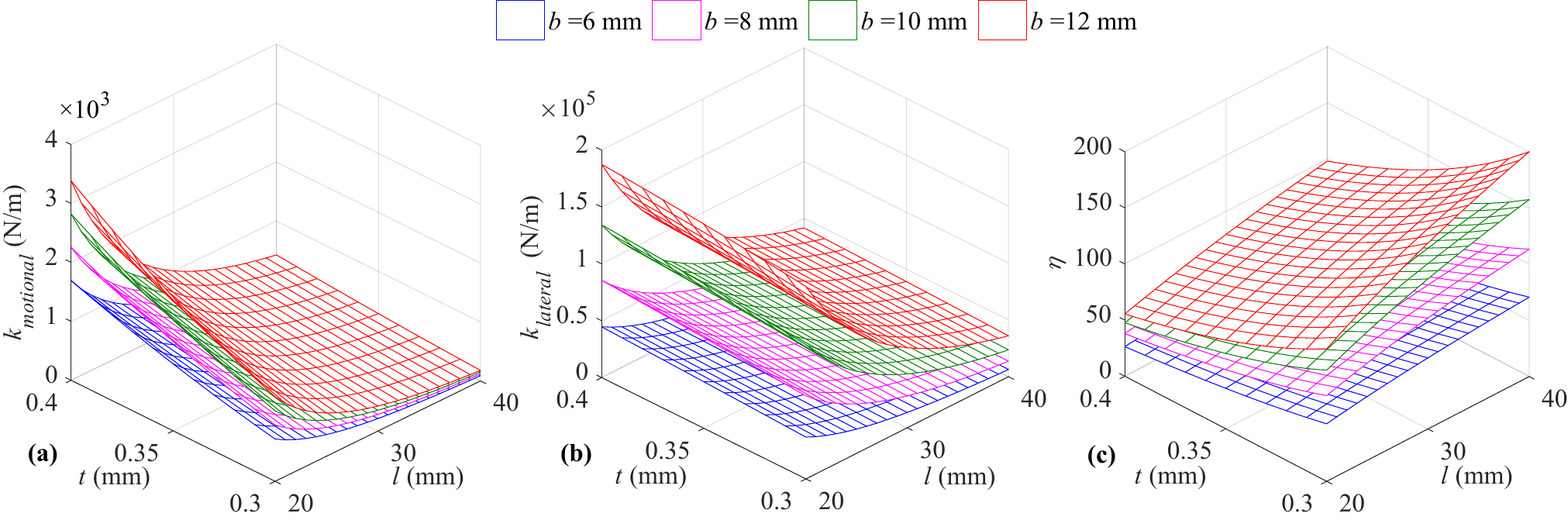}
  \caption{ Design parameter influence on (a) motional stiffness, (b) lateral stiffness and (c) stiffness ratio of MCPF.}
  \label{ParameterInfluenceOnStiffness}
\end{figure*}

For the motional stiffness of MCPF, the established model features high precision, with maximum error of 3.81\%. However, for the lateral stiffness, the maximum error reaches 12.0\%, and the error tends to increase as $l$ shortening or $b$ enlarging. This mainly results from the ignorance of the torsion effect of the compliant elements in MCPF and the nonlinearity. Because larger $b$ means more torsional strain energy \cite{connor1976analysis} and shorter $l$ aggravates the nonlinearity that is not considered in our model.
Generally, the accuracy of the established analytical model satisfies our demand, which is then applied to analyze design parameter influence on the motional stiffness, lateral stiffness and the stiffness ratio, as in Fig. \ref{ParameterInfluenceOnStiffness}. Clearly, the motional stiffness of MCPF is sensitive to thickness $t$ and length $l$, while the lateral stiffness and stiffness ratio are sensitive to length $l$ and width $b$.

\subsection{Stiffness model of the FlexDelta}
With all the MCPFs modelled properly, the motional stiffness of the entire FlexDelta can be figured out. We define $k_{xgm}^e$, $k_{xdm}^e$, $k_{zgm}^e$ and $k_{zdm}^e$ as the motional stiffness of one MCPF for XY guider, XY decoupler, Z guider and Z decoupler. Likewise, we define $k_{xgl}^e$, $k_{xdl}^e$, $k_{zgl}^e$ and $k_{zdl}^e$ the lateral stiffness of one MCPF for XY guider, XY decoupler, Z guider and Z decoupler. And the stiffness ratio of these four MCPFs are
\begin{equation}
  \label{eqStiffnessRatio}
  {{\eta }_{xg}}=\frac{k_{xgl}^{e}}{k_{xgm}^{e}},{{\eta }_{xd}}=\frac{k_{xdl}^{e}}{k_{xdm}^{e}},{{\eta }_{zg}}=\frac{k_{zgl}^{e}}{k_{zgm}^{e}},{{\eta }_{zd}}=\frac{k_{zdl}^{e}}{k_{zdm}^{e}}.
\end{equation}

\begin{figure}[htpb]
  \centering
  \includegraphics[width=8cm]{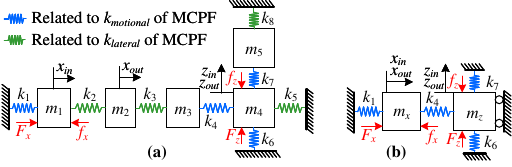}
  \caption{(a) Full and (b) simplified stiffness model of $x/y$ and $z$ branch chain.}
  \label{StiffnessModelOfBranchChain}
\end{figure}

For flexure-based parallel decoupled $xyz$ stage, its decoupling principle relies on the stiffness ratio of each MCPF which is regarded infinite by default in previous works. However, this is not the case in practice, hence the necessity to modeling the lateral stiffness of each branch chain that would be significant influenced by lateral stiffness of each MCPF. The full stiffness model of $x/y$ and $z$ branch chain is modelled as Fig.\ref{StiffnessModelOfBranchChain} (a), in which $F_x$ and $F_z$ denote the actuation force, $f_x$ and $f_z$ denote the damping force. Each spring is specified as $ {{k}_{1}}=6k_{xdm}^{e} $,  ${{k}_{2}}=2k_{zdl}^{e}$, ${{k}_{3}}=4k_{xgl}^{e}$, ${{k}_{4}}=4k_{xgm}^{e}$, ${{k}_{5}}\propto k_{zgl}^{e}$, ${{k}_{6}}=8k_{zgm}^{e}$, ${{k}_{7}}=4k_{zdm}^{e}$, ${{k}_{8}}\propto k_{xdl}^{e} $. And $m_1$ is related to the mass of XY decouplers and intermediate XY stage, while $m_2$ to output stage,  $m_3$ to XY guiders,  $m_4$ to intermediate Z stage, XY guiders and output stage,  $m_5$ to XY decouplers.

Obviously, the model is multi-DOF model for $x/y$ and $z$ axis branch chain when the lateral stiffness of MCPF is considered. Due to the existence of $k_2$, the difference between output displacement $x_{out}$ and input displacement $x_{in}$ varies inversely with $k_2$ statically and dynamically. And a smaller $k_5$ will allow undesired motion from $x/y$ to $z$ axis, likewise for smaller $k_8$ which allows motion from $z$ to $x/y$ axis. This causes cross-axis disturbance and increased coupling rate. Besides, smaller $k_2$, $k_3$ and $k_5$ will lead to lower and closer natural frequencies of each order for such multi-DOF system, making it prone to resonate.

Therefore, ensuring significantly large lateral stiffness of each MCPF is crucial to guarantee the performance of the FlexDelta. Besides, this practice helps to degrade the full models of $x/y$ and $z$ axis branch chain to single-DOF system shown as Fig.\ref{StiffnessModelOfBranchChain} (b), in which only $k_1$, $k_4$, $k_6$ and $k_7$ are considered. Given that three DOFs are kinematically decoupled, the motional stiffness of each DOF can be modelled independently. For $x$ or $y$ axis, it's
\begin{equation}
  \label{eqMotionalStiffnessx}
  {{k}_{xm}}={{k}_{1}}+{{k}_{4}}=6k_{xdm}^{e}+4k_{xgm}^{e},
\end{equation}
while for $z$ axis, it's
\begin{equation}
  \label{eqMotionalStiffnessz}
  {{k}_{zm}}={{k}_{6}}+{{k}_{7}}=8k_{zgm}^{e}+4k_{zdm}^{e}.
\end{equation}
The natural frequency of the FlexDelta is then established via lumped parameter model as 
\begin{equation}
  \label{deqn_ex9}
  \mathbf{M\ddot{X}}+\mathbf{C\dot{X}}+\mathbf{KX}=\mathbf{F},
\end{equation}
with $\mathbf{X}={{\left[ \begin{matrix}  x & y & z  \\ \end{matrix} \right]}^{\top }}$, $\mathbf{F}={{\left[ {{F}_{x}},{{F}_{y}},{{F}_{z}} \right]}^{\top }}$. And due to the assumption of fully decoupled configuration, the mass matrix, damping matrix and stiffness matrix are all diagonal, i.e. $\mathbf{M}=diag\left( {{m}_{x}},{{m}_{y}},{{m}_{z}} \right)$, $\mathbf{C}=diag\left( {{c}_{x}},{{c}_{y}},{{c}_{z}} \right)$, $\mathbf{K}=diag\left( {{k}_{xm}},{{k}_{ym}},{{k}_{zm}} \right)$. By solving determinant
\begin{equation}
  \label{eqDeterminant}
  \left| \mathbf{K}-{{\omega }^{2}}\mathbf{M} \right|=0,
\end{equation}
the natural frequency of each axis is obtainable. 

\section{Optimization Design}
Practically, the stiffness of the FlexDelta is restrained by various factors, such as the working range requirement, driving force of the motor, and natural frequency. Moreover, the stiffness ratio is also supposed significantly large to assure the dynamic performance of each branch chain and avoid coupling rate. Thus, optimization design is essential to balance these indices.

\subsection{ Optimization statement }
Generally, a larger motional stiffness is beneficial for the FlexDelta, which results in higher natural frequency. However, larger motional stiffness means stronger power demand for the driving motors, and bulky size is inevitable to safeguard huge enough stiffness ratio. Hence, a trade-off should be made among motional stiffness, stroke under limited driving force, and stiffness ratio.

Here, we restrict the maximum driving force $F_{max}$ to 50 N and require 5.5-mm unilateral stroke for sake of margin for each axis. And both $\eta_{xd}$ and $\eta_{xg}$ are empirically set larger than 60, while $\eta_{zd}$ and $\eta_{zg}$ are set higher than 100 and 20, respectively. The reason for the difference is that Z guider has small width but more quantity than Z decoupler and the former are expected to have much larger stiffness ratio in order not to cause prominent impact on motion in $xy$ plane. Further, the minimum thickness of the compliant element is constrained to 0.3 mm according to the machining capacity. Because the width $b$ of each MCPF directly influence the design process, we fix this parameter as $b_{xg}$=8 mm,  $b_{xd}$=12 mm, $b_{zd}$=6 mm, and $b_{zg}$=6 mm.

To concise the statement, optimization in $x$ or $y$ axis, and $z$ axis are separated into two statements. For $x$ or $y$ axis, the optimal design is stated as follows:

1) Objective function: maximize motional stiffness and stiffness ratio of MCPF in XY guider and XY decoupler: $\left\{ {{k}_{xm}},{{\eta }_{xd}},{{\eta }_{xg}} \right\}\to \max $.

2) Design variables:  $Par=\left[ {{t}_{xg}},{{l}_{xg}},{{t}_{xd}},{{l}_{xd}} \right]$.

3) Subject to: ${{k}_{xm}}\le {F}_{\max }/{{x}_{\max }}$, ${{\eta }_{xd}}\ge 60$, ${{\eta }_{xg}}\ge 60$, $Pa{{r}_{\min }}\in [0.3,20,0.3,30] $, and $ Pa{{r}_{\max }}\in [0.4,30,0.4,40]$.

For $z$ axis, the optimal design is stated as follows:

1) Objective function: maximize motional stiffness and stiffness ratio of MCPF in Z guider and Z decoupler:  $\left\{ {{k}_{zm}},{{\eta }_{zd}},{{\eta }_{zg}} \right\}\to \max $.

2) Design variables:  $Par=\left[ {{t}_{zg}},{{l}_{zg}},{{t}_{zd}},{{l}_{zd}} \right]$.

3) Subject to: ${{k}_{zm}}\le {F}_{\max }/{{z}_{\max }}$, ${{\eta }_{zd}}\ge 100$, ${{\eta }_{zg}}\ge 20$, $Pa{{r}_{\min }}\in [0.3,20,0.3,40] $, and $ Pa{{r}_{\max }}\in [0.4,30,0.4,50]$.
 
\subsection{Optimization results}
\begin{figure}[htpb]
  \centering
  \includegraphics[width=8cm]{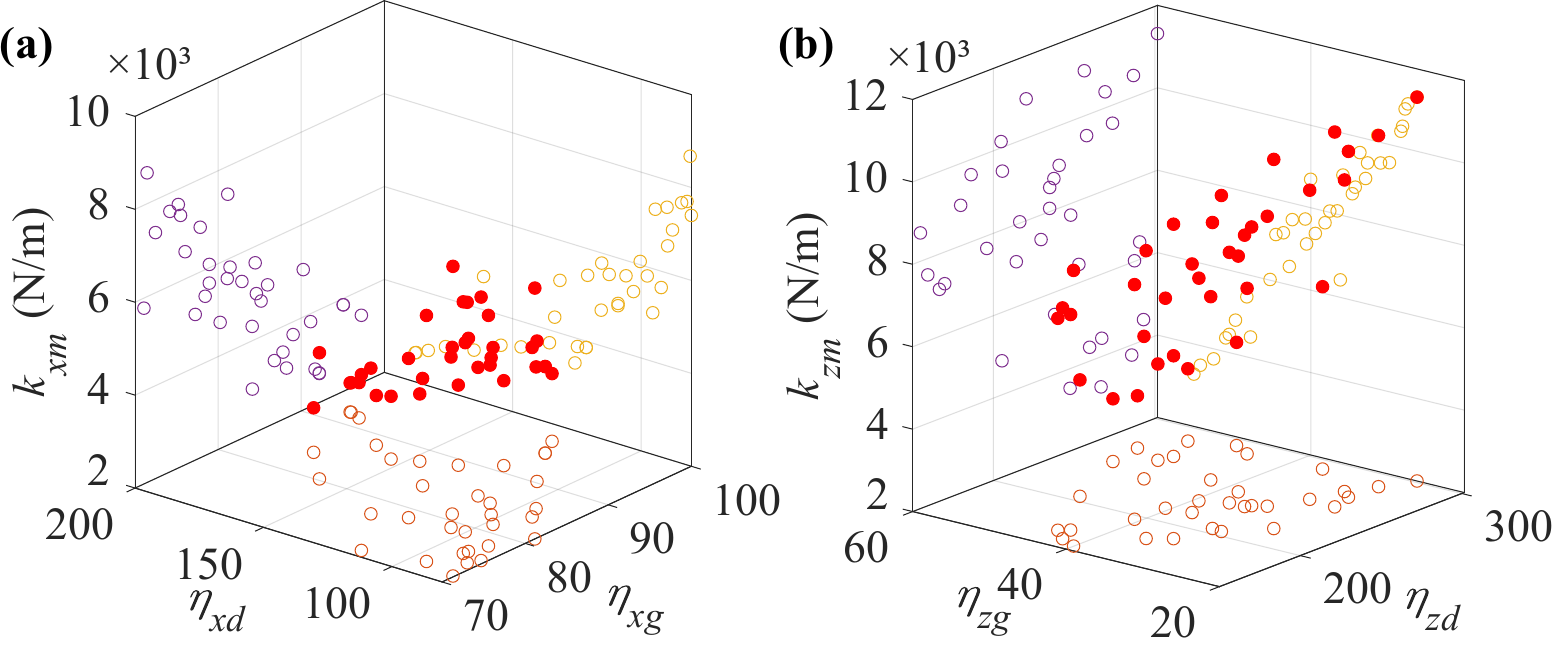}
  \caption{Optimization results of (a) $x/y$ and (b) $z$ axis.}
  \label{ParetoFront}
\end{figure}

The multi-objective optimization problem above was performed with genetic algorithm via MATLAB, in which the population size is set 200, number of generations is set 100, and mutation probability is set 0.3 for both optimal designs. Finally, 35 results are obtained for each, exhibited as Fig.\ref{ParetoFront}.

It's obvious that all these three indexes can't be optimum simultaneously for either optimal design. 
Therefore, we make a balance to choose the results which are dominated by motional stiffness requirement but still possess significantly large stiffness ratio. And we shorten their digits for machining convenience as $t_{xd}=0.40$, $l_{xd}=30.50$, $t_{xg}=0.32$,  $l_{xg}=23.00$, $t_{zd}=0.34$, $l_{zd}=22.40$, $t_{zg}=0.40$,  $l_{zg}=41.7$, all in unit of mm.
The stiffness ratio and motional stiffness of  $x/y$ and $z$ axis corresponding to the chosen optimal parameters are calculated by (\ref{eqStiffnessRatio}), (\ref{eqMotionalStiffnessx}) and (\ref{eqMotionalStiffnessz}).
And the lumped mass matrix is estimated as  $\mathbf{M}=diag\left( 0.412,0.412,0.355 \right)$, and the stiffness matrix is $\mathbf{K}=diag\left( 8787.9,8787.9,8932.1 \right)$. Consequently, the natural frequency of each axis is obtainable by solving (\ref{eqDeterminant}).
TABLE \ref{VerificationWithFEA} lists the optimal design results via analytical model.

\subsection{ Verification with FEA}
To verify the optimization results, FEA was conducted via ANSYS. TABLE \ref{VerificationWithFEA} summarizes the comparison between the optimization and FEA results. Apparently, the established model is in good conformance with FEA, with motional stiffness differing below 0.7\%, stiffness ration below 9.6\%, and natural frequency differing below 11.4\%. The negative errors of the natural frequency results from our conservative estimation of the moving mass for each axis, which is safer for early design.

\begin{table}[htpb]
  \caption{Verification with FEA}
  \centering
    \begin{threeparttable}
    \begin{tabular}{|c|| c |c c |c|c c|}
    \hline
    \multirow{2}{*}{Axis} & \multirow{2}{*}{Item}  & \multicolumn{2}{c|}{\multirow{2}{*}{Stiffness ratio}}  & Mot. Stif.  & \multicolumn{2}{c|}{Natural frequency}\\
    &  &  \multicolumn{2}{c|}{ }  & (N/m) &  \multicolumn{2}{c|}{(Hz)}  \\
    \hline
    \hline
    \multirow{4}{*}{$x/y$} &   & $\eta_{xd}$ & $\eta_{xg}$ &   $k_{xm}$   &  w/o &  w/ VCM\\
    &FEA&71.82&67.49&8832.7&28.9&25.4\\
    &Model&78.69&72.28&8787.9&25.8&23.2\\
    &Error(\%)&9.6&7.1&-0.5&-10.6&-8.6\\
    \hline
    \multirow{4}{*}{$z$} &   & $\eta_{zd}$ & $\eta_{zg}$ &   $k_{zm}$   &   &  \\
    &FEA&145.80&36.72&8994.1&32.4&28.0\\
    &Model&154.64&38.68&8932.1&28.7&25.2\\
    &Error(\%)&6.1&5.3&-0.7&-11.4&-10.0\\
    \hline
    \end{tabular}
    \label{VerificationWithFEA}
  \end{threeparttable}
\end{table}

\section{Experimental Study and Discussion}
\subsection{Experimental setup}
\begin{figure}[htpb]
  \centering
  \includegraphics[width=8cm]{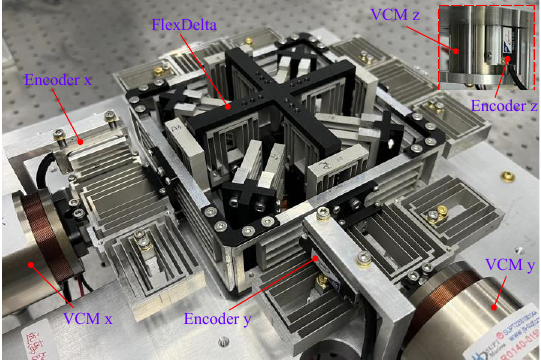}
  \caption{Experimental setup}
  \label{ExperimentalSetup}
\end{figure}

To validate the performance of proposed FlexDelta, a prototype was fabricated with Al-7075 alloy, whose flexure parts was manufactured via wire electrical discharge machining (WEDM) technique. Fig. \ref{ExperimentalSetup} illustrates the experimental setup. The FlexDelta possesses dimension of 230$\times$230$\times$60 mm$^3$ without actuators. It's actuated by three voice coil motors (VCMs), which feature 15-mm stroke and 42-N continuous thrust driven by linear drivers (TRUST Automation TA115). Three linear encoder systems (MicroE System Mercury II 5500, 100 nm) are adopted to sense the displacement of each axis. Additionally, two laser sensors (Polytec OFV-5000, 16 bits, Keyence LK-H050, 25 nm) are utilized to measure the parasitic motion.
The data acquisition task and motion control are completed with FPGA target (NI cRIO-9038) which enables high speed. Host PC with LabVIEW is employed for programming and communication with FPGA target.

\subsection{Stiffness and working space test}
By loading the FlexDelta in each axis with a screw micrometer and sensing the force with force sensor, the displacement induced is obtainable by encoders, hence the stiffness and working space.
The stroke of each axis is measured as -5.94 mm $\sim $ 6.02 mm  for $x$, -5.85 mm $\sim $ 5.86 mm  for $y$ and -4.80 mm $\sim $ 5.82 mm  for $z$ axis. Prominent asymmetry in stroke of $z$ is caused by gravity. Therefore, the working space of the FlexDelta is equivalent to 11.96$\times$11.71$\times$10.62 mm$^3$. 

\begin{figure}[htpb]
  \centering
  \includegraphics[width=8cm]{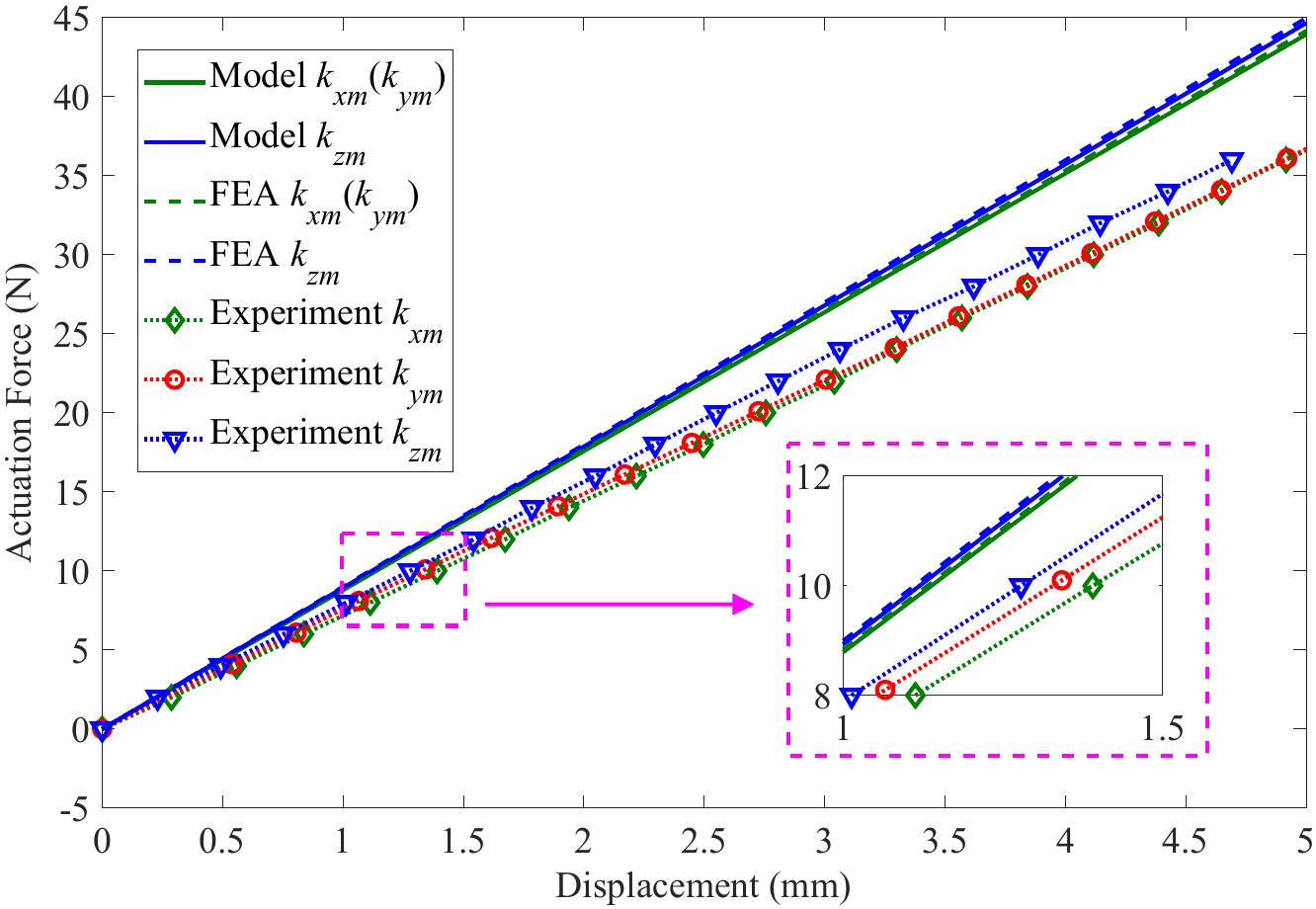}
  \caption{Results of stiffness test}
  \label{StiffnessPlot}
\end{figure}

Meanwhile, the stiffness of each axis is also tested, which is fitted to be  7339 N/m, 7299 N/m and 7677 N/m for $x$, $y$ and $z$ axis, respectively as plotted in Fig. \ref{StiffnessPlot}. Compared with FEA, the experiment results differ -16.5\%, -16.9\%, and -14.1\%. This discrepancy will be clarified in V.F.

\subsection{Coupling rate and parasitic motion test}
\begin{figure}[htpb]
  \centering
  \includegraphics[width=8cm]{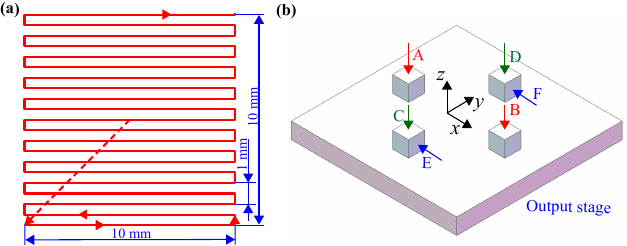}
  \caption{(a) Scanning path and (b) parasitic motion measurement schematic.}
  \label{ScanningPathAndMeasurementSchematic}
\end{figure}

To figure out the coupling rate and parasitic motion over the full range, a low-speed planar scanning path covering  10$\times$10 mm$^2$ was assigned as reference path for  $xy$, $yz$ and $xz$ plane under close-loop, as Fig. \ref{ScanningPathAndMeasurementSchematic} (a) shown. When two VCMs move to tracking the path, the third one stays unactuated and its corresponding encoder records the displacement as coupling motion. Simultaneously, two laser sensors capture the displacement difference in block face C and D on the output stage for calculating parasitic motion $\theta_x$, A and B for calculating parasitic motion $\theta_y$, E and F for calculating parasitic motion $\theta_z$, as Fig. \ref{ScanningPathAndMeasurementSchematic} (b) shown.

\begin{figure}[!h]
  \centering
  \includegraphics[width=8cm]{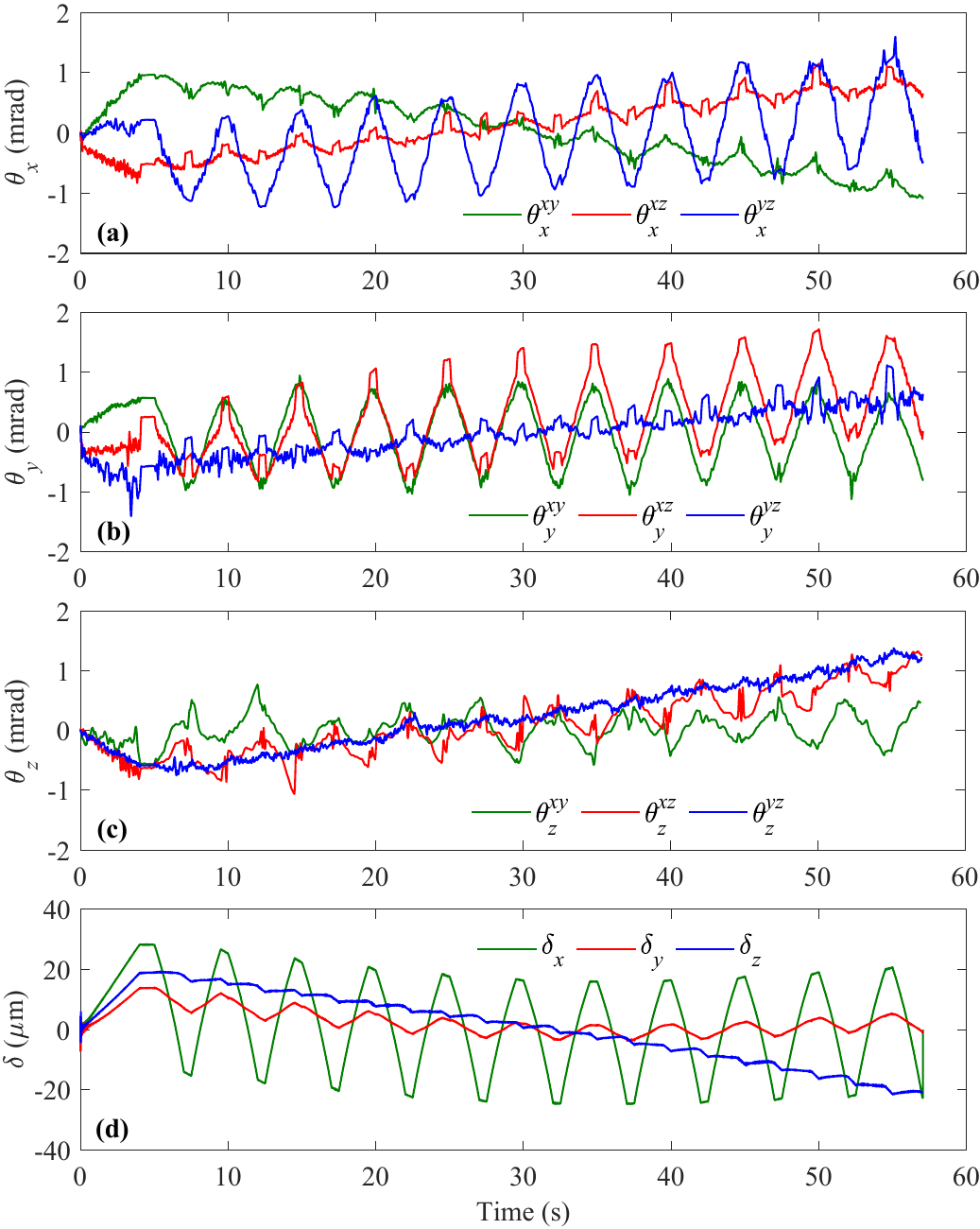}
  \caption{Parasitic motion (a) $\theta_x$, (b) $\theta_y$ and (c) $\theta_z$, and (d) coupling motion.}
  \label{CoupleRateNParasiticMotion}
\end{figure}

We define $\delta_k (k=x, y, z)$ as the coupling motion produced by other two active axes. Likewise, we denote parasitic motion as $\theta_k^{ij} ( i, j, k=x, y, z;  i \neq j $), which indicates the angular parasitic motion around $k$ axis induced by motion in $ij$ plane. With this, the coupling motion and parasitic motion are calculated, displayed as Fig. \ref{CoupleRateNParasiticMotion}. 

The maximum coupling motion for $x$, $y$, and $z$ axis are 28.3 $\mu$m, 13.9 $\mu$m, and 19.0 $\mu$m, respectively; while the minimum are -24.6 $\mu$m, -3.7 $\mu$m, and -21.4 $\mu$m individually. This corresponds to coupling rate of 0.53\%, 0.18\%, and 0.40\% over the full range for $x$, $y$, and $z$ axis, respectively.
Similarly, the maximum parasitic motion around $x$, $y$, and $z$ axis are 1.60 mrad, 1.72 mrad and 1.38 mrad separately; while the minimum are -1.22 mrad, -1.41 mrad and -1.06 mrad respectively. Therefore, the maximum coupling rate and parasitic motion over full range for FlexDelta are tested as 0.53\% and 1.72 mrad, respectively.

\subsection{System identification}
\begin{figure*}[htpb]
  \centering
  \includegraphics[width=17cm]{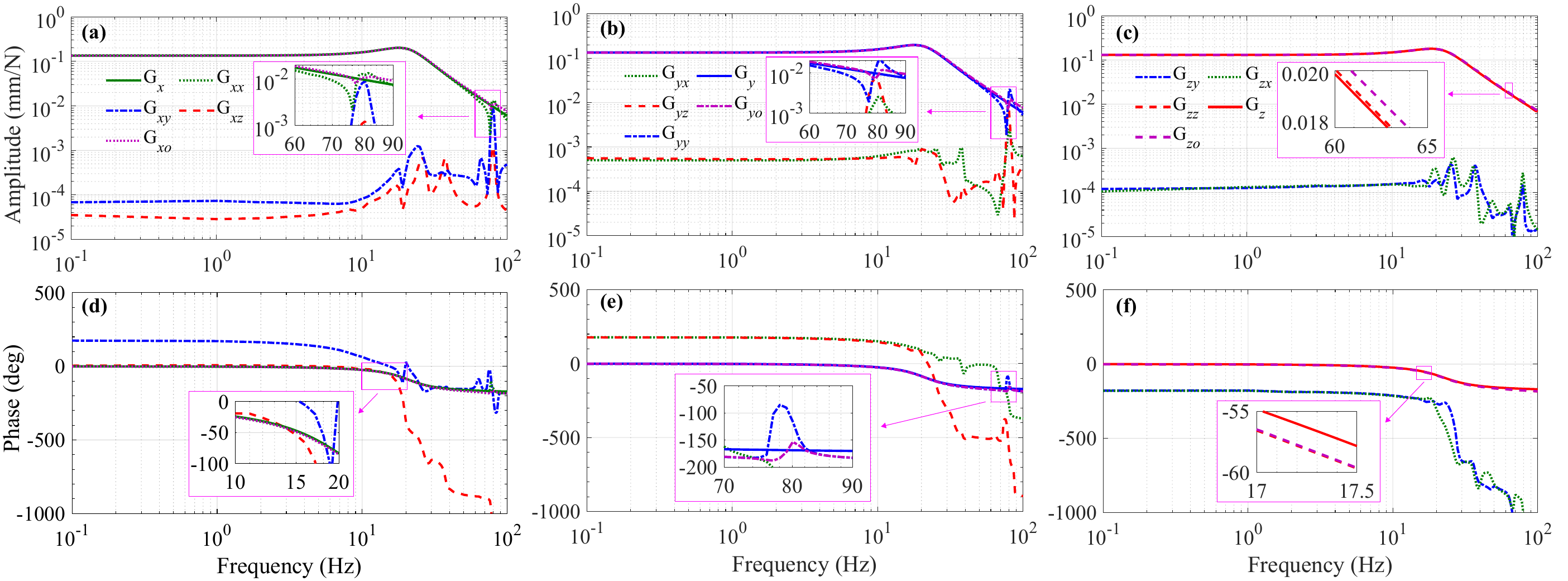}
  \caption{(a)-(c) Amplitude response and (d)-(e) phase response of $x$, $y$ and $z$.}
  \label{XYZBodePlot}
\end{figure*}

The dynamic performance of the FlexDelta under open-loop was tested by means of frequency response. A swept-sine wave with amplitude of 5 N and frequency varying from 0.1 Hz to 100 Hz was applied to VCM to stimulate each axis. Then, each encoder and a laser sensor (encoder measures input displacement and laser sensor measures output of stage) acquired their response for spectral analysis conducted via MATLAB to retrieve their amplitude and phase responses. Fig. \ref{XYZBodePlot} exhibits the system identification results.

The notation $G_{ij} (i, j = x, y, z)$ denotes the response obtained in $j$ axis when actuated in $i$ axis only, $G_{ko} (k = x, y, z)$ represents that in output stage, and $G_{k} (k = x, y, z)$ stands for the  transfer function estimated for  $G_{ij} (i=j=k)$. It's evident that the FlexDelta features excellent dynamic decoupling performance, since the amplitude in unactuated axis is hundreds of times smaller than that in dominant axis below resonant frequency. And merely minor difference between the response of input and output are seen owning to the large stiffness ratio of MCPFs. Additionally, the resonant peak and phase lag in actuated axis are minor due to the significant damping of VCM, which benefits control.

The estimated transfer function $G_x(s)$, $G_y(s)$ and $G_z(s)$ for $x$, $y$ and $z$ axis are expressed as (\ref{TransferFunction}). The natural frequency of each axis is calculated as 20.8 Hz, 20.8 Hz, and 22.4 Hz, differing -18.1\%, -18.1\%, and -20\% respectively to FEA results. This error will be explained in V.F.

\begin{equation}
  \label{TransferFunction}
  \left\{ \begin{aligned}
    & {{G}_{x}}(s)=\frac{2295}{{{s}^{2}}+92.96s+17060}({\text{mm}}/{\text{N}}\;) \\ 
   & {{G}_{y}}(s)=\frac{2280}{{{s}^{2}}+94.36s+17140}({\text{mm}}/{\text{N}}\;) \\ 
   & {{G}_{z}}(s)=\frac{2570}{{{s}^{2}}+110.9s+19760}({\text{mm}}/{\text{N}}\;) \\ 
  \end{aligned} \right.
\end{equation}

\subsection{Multi-axis path tracking performance}
In this part, 2D and 3D path tracking performance tests of the FlexDelta are carried out. Owning to the excellent decoupling performance of each axis, we adopted three classical SISO feedforward and PID feedback combined algorithm for controlling of each axis. The output of the discrete feedforward and PID feedback combined controller ($x$ axis for example) is expressed as
\begin{equation}
  \label{deqn_ex11}
  \left\{ \begin{aligned}
    & F_{c}^{k}=F_{ff}^{k}+F_{fb}^{k} \\ 
   & F_{ff}^{k}={{m}_{x}}\ddot{x}_{r}^{k}+{{c}_{x}}\dot{x}_{r}^{k}+{{k}_{xm}}x_{r}^{k} \\ 
   & F_{fb}^{k}={{k}_{p}}{{e}_{k}}+\frac{1}{{{k}_{i}}}\sum\limits_{i=0}^{k}{\frac{{{e}_{i}}+{{e}_{i-1}}}{2}}+\frac{{{k}_{d}}{{\dot{e}}_{k}}}{{{T}_{s}}+{{{k}_{d}}}/{N_f{{k}_{p}}}\;} \\ 
  \end{aligned} \right.
\end{equation}
where $k_p=150$, $k_i=0.0005$ and $k_d=0.0005$ are PID gains, $ T_s= 50 \ \mu \rm s $ is sampling time, and $N_f=50$ is the filter coefficient. The controllers for $y$ and $z$ work the same as above.

\begin{figure}[htpb]
  \centering
  \includegraphics[width=8cm]{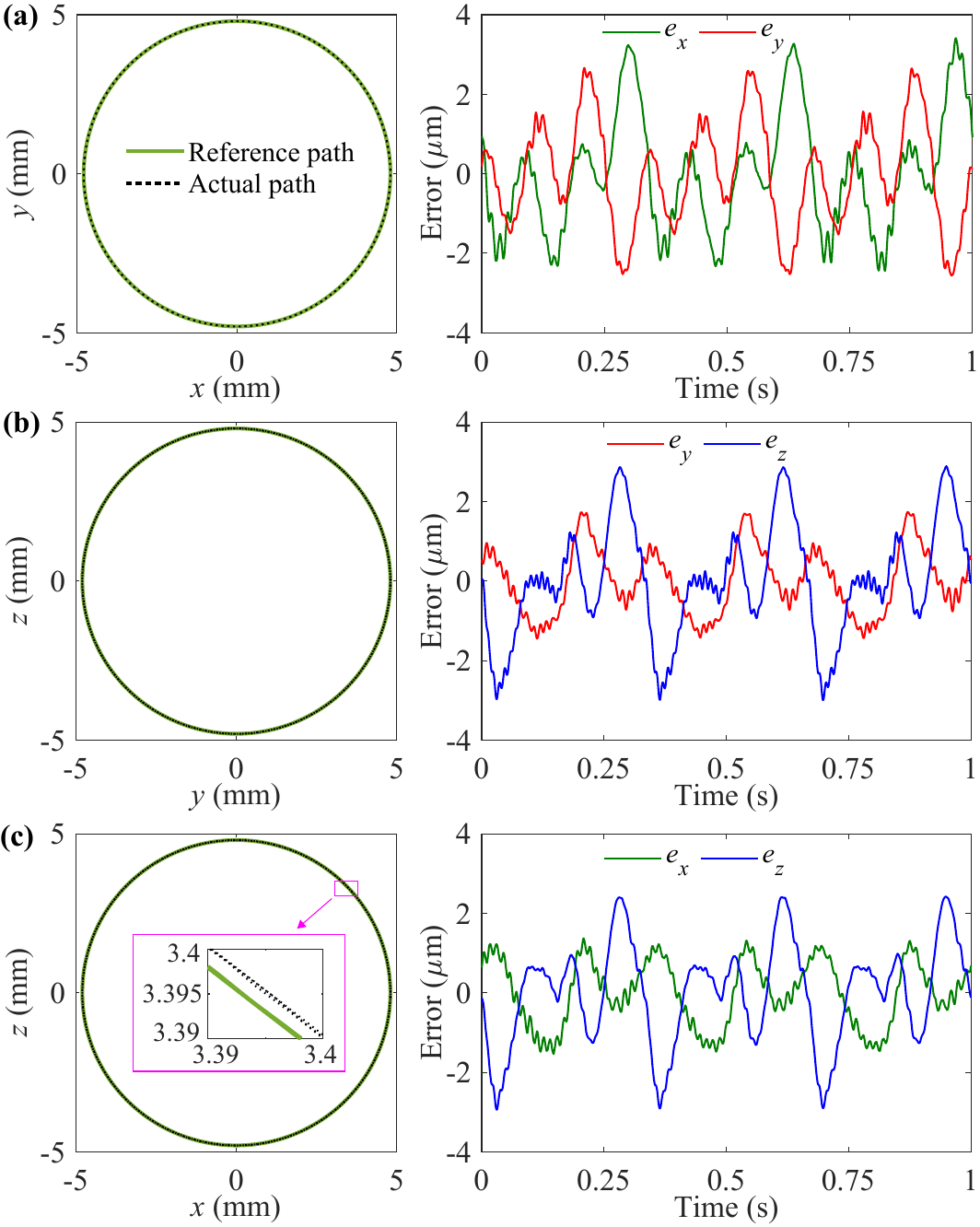}
  \caption{2D circular path tracking in (a) $xy$, (b) $yz$, (c) $xz$ plane.}
  \label{CirclePath}
\end{figure}

\begin{figure}[htpb]
  \centering
  \includegraphics[width=8cm]{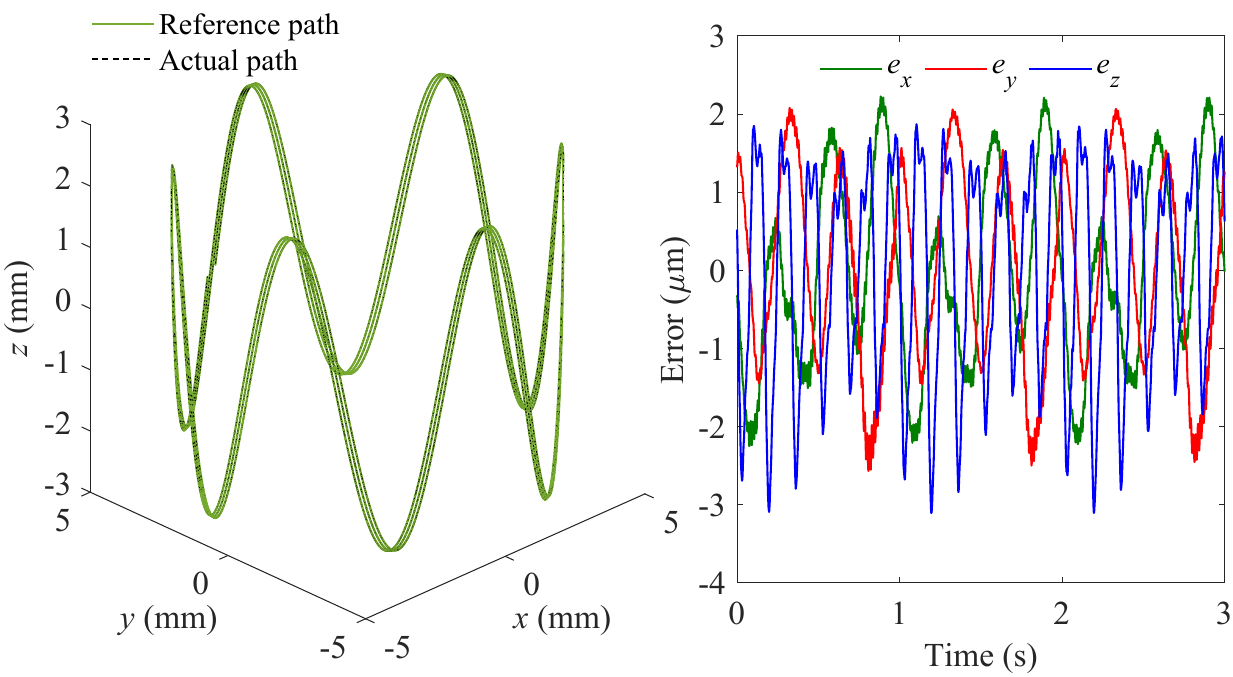}
  \caption{3D crown path tracking.}
  \label{CrownPlot}
\end{figure}

First, 2D circular reference path with diameter of 9.6 mm and travel time of 0.33 s (amplitude of 4.8 mm, frequency of 3 Hz for each component) was applied to each plane of the FlexDelta. Fig. \ref{CirclePath}  illustrates the tracking results. Then, 3D crown path (${x}_{r}=5\sin ( 2\pi t ), {y}_{r}=5\cos ( 2\pi t ), {z}_{r}=2.5\sin ( 12\pi t )$) tracking test was also carried out. Its tracking results are displayed as Fig. \ref{CrownPlot}.

TABLE \ref{TrackingError} summarizes the tracking error of both 2D and 3D path tracking errors. The multi-axis tracking results indicate the proposed FlexDelta can reach micrometer accuracy with high speed, even only under simple SISO feedforward and PID feedback combined controllers which don't take into account the dynamic couple among each axis. Therefore, it is proved the proposed FlexDelta features excellent decoupling performance both statically and dynamically.

\begin{table}[htpb]
  \caption{Multi-axis path tracking error}
  \centering
    \begin{threeparttable}
    \begin{tabular}{|c c|| c c c |c c c|}
    \hline
    \multicolumn{2}{|c||}{\multirow{2}{*}{Reference}} & \multicolumn{3}{c|}{MAXE ($\mu$m)} & \multicolumn{3}{c|}{RMSE ($\mu$m)} \\
    \cline{3-8}
    &   & $x$ & $y$ & $z$ & $x$ & $y$ & $z$ \\
    \hline
    \hline
    \multirow{2}{*}{2D}      & $xy$ & 3.41 & 2.66 & -    &  1.47 & 1.32& -    \\
    \multirow{2}{*}{circle}    & $yz$ & -    & 1.74 & 2.99 &  -    & 0.83& 1.41 \\
    \multirow{2}{*}{ }         & $xz$ & 1.54 & -    & 2.94 &  0.81 & -   & 1.28 \\
    \hline
    \multicolumn{2}{|c||}{{3D crown}} & 2.25 & 2.57 & 3.11& 1.19 &  1.20 & 1.35 \\
    \hline
    \end{tabular}
    \label{TrackingError}
  \end{threeparttable}
\end{table}

\subsection{Discussion}
\subsubsection{Performance comparison and analysis}
The decouple parallel $ xyz $ stage with long stroke and high-speed positioning capacity has been researched for decades. However, due to its complicated configuration, it has been challenging to satisfy various indexes simultaneously. Here, TABLE \ref{PerformanceComparison} lists some typical relevant works and makes comparison with proposed FlexDelta. It's evident that the proposed design outstands in most critical indexes, especially in parasitic motion and coupling rate. And it possesses largest compactness ratio, which is defined as the ratio between working space and dimension, compared with other relevant works to our best knowledge. The natural frequency of this work is also remarkable compared with those that have similar working space. These characteristics are of great significance for positioning applications where both high speed and precision are required.

\begin{table*}[htpb]
  \caption{Performance comparison of decouple parallel $xyz$ stage with long stroke (above millimeter)}
  \centering
    \begin{threeparttable}
    \begin{tabular}{|c||c|c|c|c|c|c|}
    \hline
    \multirow{2}{*}{Reference} &Working space  & Dimension  & Compactness ratio  & Natural frequency\tnote{d}  & Parasitic motion  & Coupling rate\\
    & (mm$^3$) & (mm$^3$) & (\textperthousand) & (Hz) & (mrad) & (\%) \\
    \hline
    \hline
    \cite{tang2006large} & 1$\times$1$\times$1 & 300$\times$150$\times$200 & \textless 0.001 & 56 & 1.5 ($\theta_z$) & 1.9 ($\delta_z$)\\

    \cite{hao2015design} & 1.5$\times$1.5$\times$1.5 & 105$\times$105$\times$105 & 0.003 & 182.7 & - & -\\

    \cite{bacher2001delta3} & 2$\times$2$\times$2 & 100$\times$100$\times$100 & 0.008 & 8.5 & - & -\\

    \cite{zhang2018design} & 2.2$\times$2.2$\times$1.8 & 176$\times$176$\times$198 & 0.001 & 14 (FEA:17.8) & - & 1.7\\

    \cite{hao20093}\tnote{a} & 9.5$\times$9.5$\times$9.5 & 300$\times$300$\times$300 & 0.032 & (FEA:9.0) & - & 1\\

    \cite{awtar2021experimental} & 10$\times$10$\times$10 & 150$\times$150$\times$150 & 0.296 & (FEA:27.6)\tnote{b} & 9.5 (F.R.)\tnote{c} & 11.6 (F.R.)\tnote{c}\\

    This work & 11.9$\times$11.7$\times$10.6 & 230$\times$230$\times$60 & 0.465 & 20.8 (FEA:25.4) & 1.72 (F.R.)\tnote{c}& 0.53 (F.R.)\tnote{c}\\
    \hline
    \end{tabular}
    \label{PerformanceComparison}
    \begin{tablenotes}
      \footnotesize
      \item [a] This work provides no experimental study.
      \item [b] Not provided in original paper. Simulated by authours according to the dimension of mechanism (without actuator) presented, for reference only.
      \item [c] The parasitic motion and coupling rate in these two works are measured over the full range (F.R.), which indicate the largest value within working space.
      \item [d] The smallest one among three axes.
    \end{tablenotes}
  \end{threeparttable}
\end{table*}

We attribute the favorable performance of this work to two aspects. The first one is the design aspect, i.e., symmetric design of each axis, coaxial arrangement of parts consisting each axis as elaborated in Section II, the adoption of MCPF, as well as the nested design of Z and XY guiders that enhances the space utilization. The former three points essentially alleviate the parasitic motion and coupling rate even over long stroke, particularly the symmetric design which is neglected in almost all other works listed in TABLE \ref{PerformanceComparison}. 
The other aspect is the consideration of lateral stiffness during modelling and optimization design. The motion decouple between $x/y$ and $z$ relies on large stiffness ratio of related flexures, otherwise cross-axis disturbance and coupling motion will be inevitable. Besides, due to its existence, the stiffness model of each branch chain is actually multi-DOF system, where the input and output displacement occur at different position for $x/y$ axis. This leads to the motion difference between input and output. By guaranteeing large stiffness ratio of flexure in optimization design, these problems can be mitigated.

Another source of coupling motion comes from the axial stiffness (along the length) of flexure hinges, which varies with motion in other axes. This is not considered in this work, because it is a principle flaw of all kinds of parallelogram-based flexure and can hardly be mitigated by parameter optimization design. Concrete analysis of this phenomenon has been presented in \cite{awtar2007characteristics}. Despite this, the remaining coupling rate after careful design and optimization through lateral stiffness is less than 30 $\mu$m from the experiment results, which can be well compensated by close-loop control as the multi-axis path tracking in V.E shown.

\subsubsection{Discrepancy between experiment and FEA}
Next, we'd like to analyze the sources of discrepancy between experiment and FEA in motional stiffness and natural frequency mentioned in V.B and V.D. The main reason is blame to the machining error. Because the thinnest thickness of all flexure, to which the motional stiffness and natural frequency is sensitive, is within 0.4 mm, making it hard to guarantee for affordable WEDM. A brief analysis is given as follows.

Let $k_{xm}^{n}$ = 8832.7 N/m, $k_{ym}^{n}$ = 8832.7 N/m, $k_{zm}^{n}$ = 8994.1 N/m be the nominal motional stiffness of $x$, $y$, and $z$ axis. And $k_{xm}^{a}$ = 7339 N/m, $k_{ym}^{a}$ = 7299 N/m, $k_{zm}^{a}$ = 7677 N/m denote the actual motional stiffness of each axis, respectively. Then the nominal motional stiffness and actual one of each axis has following relation
\begin{equation}
  k_{xm}^{a}=\alpha_x k_{xm}^{n}, k_{ym}^{a}=\alpha_y k_{ym}^{n}, k_{zm}^{a}=\alpha_z k_{zm}^{n}
\end{equation}
where $\alpha_x$=0.831, $\alpha_y$=0.818, $\alpha_z$=0.859 denote proportional coefficient between nominal motional stiffness and actual one for each axis. According to \cite{xu2013design}, the motional stiffness is approximately proportional to $b_{eq}t_{eq}^3/l_{eq}^3$, in which the equivalent length $l_{eq}$ and width $b_{eq}$ of flexure is easier to machined accurately, while the equivalent thichness $t_{eq}$ is the most difficult parameter to guarantee via WEDM technique due to its small dimension. 

By assuming accurate machining of $l_{eq}$ and $b_{eq}$, it is estimated that the machining errors for $t_{eq}$ are -6.0\%, -6.5\% and -4.9\% for $x$, $y$, and $z$ axis, respectively. This can be very reasonable for ordinary affordable WEDM technique.
Based on this inference, the corrected actual natural frequency for $x$, $y$, and $z$ axis is
\begin{equation}
  \left\{ \begin{aligned}
    \hat f_{nx}^{a}=\sqrt{\alpha_x } f_{nx}^{n} =23.2 \ \rm{Hz} \\
    \hat f_{ny}^{a}=\sqrt{\alpha_y } f_{ny}^{n}=23.0 \ \rm{Hz} \\
    \hat f_{nz}^{a}=\sqrt{\alpha_z } f_{nz}^{n} =26.0 \ \rm{Hz} 
  \end{aligned} \right.
\end{equation}
to which the actual natural frequency $ f_{nx}^{a}$ = $f_{ny}^{a}$ = 20.8 Hz, $f_{nz}^{a}$=22.4 Hz differ -10.1\%, -9.6\% and -13.8\%. This makes sense considering the model used for FEA and actual prototype is not totally identical. One can tell from Fig. \ref{DesignOverviewOfTheProposedFlexDelta} and Fig. \ref{ExperimentalSetup} that modification is made on the model. This is essential to facilitate its manufacturing and assembly, although it in turn leads to extra parts and considerable screws ignored in FEA that increase mass thus decreasing the natural frequency.

\subsubsection{Manufacture and assembly consideration}
The prototype in this work is manufactured part by part then assembly together for sake of reducing machining demand. Because multi-DOF positioning stages are typically complicated thus inappropriate to manufacture monolithically so far like single-DOF or dual-DOF ones. However, unlike conventional roller bearing-based multi-DOF stages which require accurate assembling technique due to massive motion pairs, the assembly of flexure-based positioning stages only includes bolt connection. This is far less demanding and can be accomplished even by inexperienced practitioners. The most critical point for manufacturing such complicated positioning stage is still the precise machining of flexure, especially its thickness, as discussed before.

\section{Conclusion}
The conceptual design, modeling, and experimental study of a novel decoupled parallel $xyz$ positioning stage based on flexure guides (FlexDelta) is presented in this paper. Firstly, the working principle of FlexDelta is introduced, followed by its mechanism design with flexure. Secondly, the stiffness model of flexure is established via matrix-based Castigliano's second theorem, where the influence of its lateral stiffness on the stiffness model of mechanism is comprehensively investigated and then optimally designed. Finally, a prototype was fabricated based on which experimental study was implemented. The results reveal that the positioning stage features centimeter-stroke in three axes, with coupling rate less than 0.53\%, parasitic motion less than 1.72 mrad over full range. And the natural frequencies are 20.8 Hz, 20.8 Hz, and 22.4 Hz for $x$, $y$, and $z$ axis respectively. Multi-axis path tracking tests were also carried out, which validates its dynamic performance with micrometer error. And some critical points involving its performance, errors,manufacture and assembly consideration are exhaustively discussed.
Further effort in accuracy improvement will be considered in our future work.

\bibliographystyle{Bibliography/IEEEtran}
\bibliography{Bibliography/RefDecoupledParallelXYZ}

\begin{thebibliography}{10}
\providecommand{\url}[1]{#1}
\csname url@samestyle\endcsname
\providecommand{\newblock}{\relax}
\providecommand{\bibinfo}[2]{#2}
\providecommand{\BIBentrySTDinterwordspacing}{\spaceskip=0pt\relax}
\providecommand{\BIBentryALTinterwordstretchfactor}{4}
\providecommand{\BIBentryALTinterwordspacing}{\spaceskip=\fontdimen2\font plus
\BIBentryALTinterwordstretchfactor\fontdimen3\font minus
  \fontdimen4\font\relax}
\providecommand{\BIBforeignlanguage}[2]{{%
\expandafter\ifx\csname l@#1\endcsname\relax
\typeout{** WARNING: IEEEtran.bst: No hyphenation pattern has been}%
\typeout{** loaded for the language `#1'. Using the pattern for}%
\typeout{** the default language instead.}%
\else
\language=\csname l@#1\endcsname
\fi
#2}}
\providecommand{\BIBdecl}{\relax}
\BIBdecl

\bibitem{lafratta2020making}
C.~N. LaFratta and L.~Li, ``Making two-photon polymerization faster,'' in
  \emph{Three-Dimensional Microfabrication Using Two-Photon
  Polymerization}.\hskip 1em plus 0.5em minus 0.4em\relax Elsevier, 2020, pp.
  385--408.

\bibitem{serge2020motion}
M.~Serge, T.~Patrick, F.~Duquenoy, and P.~N. Dinh, ``Motion systems: An
  overview of linear, air bearing, and piezo stages,'' \emph{Three-Dimensional
  Microfabrication Using Two-photon Polymerization}, pp. 303--325, 2020.

\bibitem{zhang2017large}
Z.~Zhang, P.~Yan, and G.~Hao, ``A large range flexure-based servo system
  supporting precision additive manufacturing,'' \emph{Engineering}, vol.~3,
  no.~5, pp. 708--715, 2017.

\bibitem{zhang2019robotic}
Z.~Zhang, X.~Wang, J.~Liu, C.~Dai, and Y.~Sun, ``Robotic micromanipulation:
  Fundamentals and applications,'' \emph{Annual Review of Control, Robotics,
  and Autonomous Systems}, vol.~2, pp. 181--203, 2019.

\bibitem{werner2010design}
C.~Werner, P.~Rosielle, and M.~Steinbuch, ``Design of a long stroke translation
  stage for afm,'' \emph{International journal of machine tools and
  manufacture}, vol.~50, no.~2, pp. 183--190, 2010.

\bibitem{weckenmann2007long}
A.~Weckenmann and J.~Hoffmann, ``Long range 3 d scanning tunnelling
  microscopy,'' \emph{CIRP annals}, vol.~56, no.~1, pp. 525--528, 2007.

\bibitem{bettahar20206}
H.~Bettahar, O.~Lehmann, C.~Cl{\'e}vy, N.~Courjal, and P.~Lutz, ``6-dof full
  robotic calibration based on 1-d interferometric measurements for microscale
  and nanoscale applications,'' \emph{IEEE Transactions on Automation Science
  and Engineering}, 2020.

\bibitem{zhan2018error}
Z.~Zhan, X.~Zhang, Z.~Jian, and H.~Zhang, ``Error modelling and motion
  reliability analysis of a planar parallel manipulator with multiple
  uncertainties,'' \emph{Mechanism and Machine Theory}, vol. 124, pp. 55--72,
  2018.

\bibitem{yang2022along}
M.~Yang, C.~Zhang, X.~Huang, S.~Chen, and G.~Yang, ``A long-stroke
  nanopositioning stage with annular flexure guides,'' \emph{IEEE/ASME
  Transactions on Mechatronics}, vol.~27, no.~3, pp. 1570--1581, 2022.

\bibitem{du2013piezo}
Z.~Du, R.~Shi, and W.~Dong, ``A piezo-actuated high-precision flexible parallel
  pointing mechanism: conceptual design, development, and experiments,''
  \emph{IEEE transactions on robotics}, vol.~30, no.~1, pp. 131--137, 2013.

\bibitem{chen2019pzt}
F.~Chen, W.~Dong, M.~Yang, L.~Sun, and Z.~Du, ``A pzt actuated 6-dof
  positioning system for space optics alignment,'' \emph{IEEE/ASME Transactions
  on Mechatronics}, vol.~24, no.~6, pp. 2827--2838, 2019.

\bibitem{hesselbach2004performance}
J.~Hesselbach, A.~Raatz, and H.~Kunzmann, ``Performance of pseudo-elastic
  flexure hinges in parallel robots for micro-assembly tasks,'' \emph{CIRP
  Annals}, vol.~53, no.~1, pp. 329--332, 2004.

\bibitem{xie2021design}
Y.~Xie, Y.~Li, C.~F. Cheung, Z.~Zhu, and X.~Chen, ``Design and analysis of a
  novel compact xyz parallel precision positioning stage,'' \emph{Microsystem
  Technologies}, vol.~27, no.~5, pp. 1925--1932, 2021.

\bibitem{yun2011optimal}
Y.~Yun and Y.~Li, ``Optimal design of a 3-pupu parallel robot with compliant
  hinges for micromanipulation in a cubic workspace,'' \emph{Robotics and
  Computer-Integrated Manufacturing}, vol.~27, no.~6, pp. 977--985, 2011.

\bibitem{bacher2001delta3}
J.-P. Bacher, S.~Bottinelli, J.-M. Breguet, and R.~Clavel, ``Delta3: design and
  control of a flexure hinge mechanism,'' in \emph{Microrobotics and
  microassembly III}, vol. 4568.\hskip 1em plus 0.5em minus 0.4em\relax
  International Society for Optics and Photonics, 2001, pp. 135--142.

\bibitem{tang2006large}
X.~Tang and I.-M. Chen, ``A large-displacement 3-dof flexure parallel mechanism
  with decoupled kinematics structure,'' in \emph{2006 IEEE/RSJ International
  Conference on Intelligent Robots and Systems}.\hskip 1em plus 0.5em minus
  0.4em\relax IEEE, 2006, pp. 1668--1673.

\bibitem{li2010totally}
Y.~Li and Q.~Xu, ``A totally decoupled piezo-driven xyz flexure parallel
  micropositioning stage for micro/nanomanipulation,'' \emph{IEEE Transactions
  on Automation Science and Engineering}, vol.~8, no.~2, pp. 265--279, 2010.

\bibitem{hao2015design}
G.~Hao and H.~Li, ``Design of 3-legged xyz compliant parallel manipulators with
  minimised parasitic rotations,'' \emph{Robotica}, vol.~33, no.~4, pp.
  787--806, 2015.

\bibitem{hao20093}
G.~Hao and X.~Kong, ``A 3-dof translational compliant parallel manipulator
  based on flexure motion,'' in \emph{International Design Engineering
  Technical Conferences and Computers and Information in Engineering
  Conference}, vol. 49040, 2009, pp. 101--110.

\bibitem{awtar2021experimental}
S.~Awtar, J.~Quint, and J.~Ustick, ``Experimental characterization of a
  large-range parallel kinematic xyz flexure mechanism,'' \emph{Journal of
  Mechanisms and Robotics}, vol.~13, no.~1, 2021.

\bibitem{zhang2018design}
X.~Zhang and Q.~Xu, ``Design, fabrication and testing of a novel symmetrical
  3-dof large-stroke parallel micro/nano-positioning stage,'' \emph{Robotics
  and Computer-Integrated Manufacturing}, vol.~54, pp. 162--172, 2018.

\bibitem{chen2022design}
X.~Chen, Y.~Li, Y.~Xie, and R.~Wang, ``Design and analysis of new ultra compact
  decoupled xyz$\theta$ stage to achieve large-scale high precision motion,''
  \emph{Mechanism and Machine Theory}, vol. 167, p. 104527, 2022.

\bibitem{chen2021design}
Z.~Chen, J.~Shi, S.~Zhu, X.~Zhong, and X.~Zhang, ``Design and testing of a
  damped piezo-driven decoupled xyz stage,'' in \emph{2021 IEEE International
  Conference on Robotics and Automation (ICRA)}.\hskip 1em plus 0.5em minus
  0.4em\relax IEEE, 2021, pp. 6986--6991.

\bibitem{lin2019decoupling}
C.~Lin, J.~Yu, Z.~Wu, and Z.~Shen, ``Decoupling and control of micromotion
  stage based on hysteresis of piezoelectric actuation,'' \emph{Microsystem
  Technologies}, vol.~25, pp. 3299--3309, 2019.

\bibitem{ling2018design}
M.~Ling, J.~Cao, Q.~Li, and J.~Zhuang, ``Design, pseudostatic model, and
  pvdf-based motion sensing of a piezo-actuated xyz flexure manipulator,''
  \emph{IEEE/ASME Transactions on Mechatronics}, vol.~23, no.~6, pp.
  2837--2848, 2018.

\bibitem{liu2018review}
Y.~Liu, J.~Deng, and Q.~Su, ``Review on multi-degree-of-freedom piezoelectric
  motion stage,'' \emph{IEEE Access}, vol.~6, pp. 59\,986--60\,004, 2018.

\bibitem{zhang20223}
S.~Zhang, H.~Zhao, X.~Ma, J.~Deng, and Y.~Liu, ``A 3-dof piezoelectric
  micromanipulator based on symmetric and antisymmetric bending of a
  cross-shaped beam,'' \emph{IEEE Transactions on Industrial Electronics},
  2022.

\bibitem{xu2011new}
Q.~Xu, ``New flexure parallel-kinematic micropositioning system with large
  workspace,'' \emph{IEEE Transactions on Robotics}, vol.~28, no.~2, pp.
  478--491, 2011.

\bibitem{xu2013design}
------, ``Design and development of a compact flexure-based $ xy $ precision
  positioning system with centimeter range,'' \emph{IEEE Transactions on
  Industrial Electronics}, vol.~61, no.~2, pp. 893--903, 2013.

\bibitem{connor1976analysis}
J.~J. Connor, \emph{Analysis of structural member systems}.\hskip 1em plus
  0.5em minus 0.4em\relax Ronald Press Company, 1976.

\bibitem{awtar2007characteristics}
S.~Awtar, A.~H. Slocum, and E.~Sevincer, ``Characteristics of beam-based
  flexure modules,'' \emph{Journal of Mechanical Design}, vol. 129, p. 625,
  2007.

\end{thebibliography}

\end{document}